\documentclass{article}

\usepackage{microtype}
\usepackage{graphicx}
\usepackage{subfigure}
\usepackage{booktabs} % for professional tables

\usepackage{hyperref}

\usepackage[accepted]{icml2023}

\usepackage{amsmath}
\usepackage{amssymb}
\usepackage{mathtools}
\usepackage{amsthm}

\usepackage[capitalize,noabbrev]{cleveref}

\usepackage{amsmath,amssymb,amsfonts,amsthm}
\usepackage{graphicx}
\usepackage{textcomp}
\usepackage{xcolor}
\usepackage{nicefrac,enumerate}
\usepackage{tcolorbox}
\tcbset{width = \columnwidth, center,before skip=20pt,after skip=15pt}

 \newtheorem*{theorem*}{Theorem}
 \newtheorem*{lemma*}{Lemma}
 \newtheorem{ex}{Example}

\newcommand{\relu}{\text{ReLU}}

\newcommand{\I}{I_x^{\alpha}}
\newcommand{\Iy}{I_y^{\alpha}}

\newcommand{\xx}{\langle x,x_i\rangle}
\newcommand{\yx}{\langle y,x_i\rangle}

\newcommand{\Ta}{C_\alpha}
\newcommand{\RR}{\mathbb{R}}
\newcommand{\Sp}{\mathbb{S}}

\theoremstyle{plain}
\newtheorem{theorem}{Theorem}[section]

\newtheorem{lemma}[theorem]{Lemma}

\theoremstyle{definition}
\newtheorem{definition}[theorem]{Definition}

\theoremstyle{remark}
\newtheorem{remark}[theorem]{Remark}

\usepackage[textsize=tiny]{todonotes}

\icmltitlerunning{Injectivity and Local Reconstruction of ReLU-layers}

\definecolor{darkviolet}{rgb}{0.58,0,0.83} %{148,0,211}

\begin{document}

\twocolumn[
\icmltitle{Convex Geometry of ReLU-Layers,\\Injectivity on the Ball and Local Reconstruction}




\begin{icmlauthorlist}
\icmlauthor{Daniel Haider}{yyy,comp}
\icmlauthor{Martin Ehler}{comp}
\icmlauthor{Peter Balazs}{yyy}
\end{icmlauthorlist}

\icmlaffiliation{yyy}{Acoustics Research Institute, Vienna, Austria}
\icmlaffiliation{comp}{University of Vienna, Department of Mathematics, Austria}

\icmlcorrespondingauthor{Daniel Haider}{daniel.haider@oeaw.ac.at}
%\icmlcorrespondingauthor{Firstname2 Lastname2}{first2.last2@www.uk}

\icmlkeywords{ReLU, frames, injectivity, reconstruction, convex polytopes}

\vskip 0.3in]



\printAffiliationsAndNotice{}  % leave blank if no need to mention equal contribution

\begin{abstract}
The paper uses a frame-theoretic setting to study the injectivity of a ReLU-layer on the closed ball of $\RR^n$ and its non-negative part.
In particular, the interplay between the radius of the ball and the bias vector is emphasized. Together with a perspective from convex geometry, this leads to a computationally feasible method of verifying the injectivity of a ReLU-layer under reasonable restrictions in terms of an upper bound of the bias vector. Explicit reconstruction formulas are provided, inspired by the duality concept from frame theory. All this gives rise to the possibility of quantifying the invertibility of a ReLU-layer and a concrete reconstruction algorithm for any input vector on the ball. 
\end{abstract}

\section{Introduction} 
The Rectified Linear Unit $\relu(s) = \max(0,s)$, $s\in \RR$ has become indispensable in modern neural network architecture. It is applied component-wise on the output of an affine linear function $Ax-b$, comprising of the multiplication by a weight matrix $A$ and the shift by a bias vector $b$. The combined mapping is called a \emph{ReLU-layer}. This has proven to be a simple, yet effective non-linear mapping to handle fundamental problems in the training of deep neural networks well \cite{deepsparse11,hin12,dlb,relu10}. Despite its simplicity, yet, the ReLU function still hides some mysteries and is an active topic of research \cite{dittmer2018singrelu}.

Recently, invertible network architectures have been getting a lot of attention due to their increased interpretability and the possibility of reversing the forward process analytically, which is especially interesting in a generative setting. This found many applications in the context of normalizing flows, offering exact and efficient likelihood estimations \cite{normflows17,adv17}. Mathematically speaking, the forward process in such an invertible architecture must be \textit{injective}, guaranteeing the existence of a \textit{left-inverse} that allows perfect reconstruction of any input. A ReLU-layer is a mapping that is designed to provide sparse output. Hence, its injectivity is an interesting property that has been tackled from a theoretical point of view only little in the literature. Bruna et al. characterized a ReLU-layer to be injective in terms of an admissibility condition for index sets and proved a bi-Lipschitz stability condition for an injective ReLU-layer, see Proposition 2.2 in \cite{bruna14}. Just recently, Puthawala et al. formulated a condition in terms of spanning sets that is equivalent to the one in \cite{bruna14} (with a slight modification) and describes the injectivity of ReLU-networks consisting of many ReLU-layers see Theorem $2$ in \cite{puth22}. Both conditions, however, are not applicable to verify the injectivity of a ReLU-layer for a given weight matrix in practice. The presented work provides exactly that. We found the convex geometry of the weight matrix to play an essential role in the injectivity analysis for the associated ReLU-layer, using a concept that Behrmann et al. introduced in Theorem 4 of \cite{behr18}. The geometrical perspective helps profoundly to strengthen the intuition on the effect of the ReLU function. It allows to formulate a computationally feasible method to give a sufficient condition for injectivity. 
This shall contribute to the enhancement of the interpretability of neural networks in terms of a way to quantify the invertibility of a ReLU-layer with corresponding exact reconstruction formulas. Aiming to set a rigorous foundation for future work on this topic, we formulate all results in an abstract mathematical manner, using the language of \textit{frame theory} which we find to be especially well-suited.

In Section 2 we interpret a ReLU-layer by means of frame theory and motivate the restriction to the ball. Section 3 is dedicated to the injectivity of a ReLU-layer theoretically. In Section 4 we introduce a method to obtain an upper bound for all biases, such that the corresponding ReLU-layer is injective on the ball and its non-negative part. Explicit reconstruction formulas are stated. Finally, Section 5 demonstrates how the method can be used to analyze the injectivity behavior of a ReLU-layer in numerical experiments.

\section{Mathematical Context}
\subsection{Neural Networks meet Frame Theory}
The goal of this section is to link abstract frame theory with deep learning. We want to particularly emphasize that frames are a well-suited concept for the mathematical analysis of neural networks, not only in terms of notation but also due to its long usage in signal processing which is tied closely to deep learning. In this sense, we build our work upon notation and tools from frame theory for $\RR^n$, c.f. \cite{xxlfinfram1,casfin12}.
We shall write
$$
X=(x_i)_{i\in I}\subseteq \mathbb{R}^n \quad \text{ with } \quad \vert I \vert=m\geq n
$$
to refer to a collection of $m$ vectors $x_1,\ldots,x_m$ in $\RR^n$. Denoting the usual inner product on $\RR^n$ as $\langle \cdot,\cdot \rangle $ we say that $X$ constitutes a \emph{frame} for $\RR^n$ with \textit{frame elements} $x_i$, if there are constants $0<A\leq B<\infty$, such that
\begin{equation}\label{eq:frame}
    A\cdot \|x\|^2 \leq \sum_{i\in I} \vert \xx \vert^2 \leq B\cdot \|x\|^2
\end{equation}
holds for all $x\in \RR^n$. The constants $A, B$ are called lower and upper frame bounds for $X$.
% {\color{red} 
In $\RR^n$, a frame is equivalent to a spanning set.
% (In infinite-dimensional Hilbert spaces, \eqref{eq:frame} becomes even more crucial.)
% In the course of stability analysis within the paper, also the general definition of a frame in terms of upper and lower frame bounds will be of importance: $X$ is a frame for $\RR^n$ if there are constants $0< A\leq B<\infty$ such that
% \begin{equation}\label{eq:framedef}
%     A\cdot \|x\|^2\leq \sum_{i\in I}\vert \xx \vert^2 \leq B\cdot\| x\|^2
% \end{equation}
% holds for all $x\in\mathbb{R}^n$. Clearly, the upper inequality always holds true in finite dimensions and the lower one is equivalent to the spanning property. 
% Nevertheless, t
The bounds $A, B$ become important, if one is interested in the numerical properties of the operators associated with a frame: the \emph{analysis operator}
\begin{align*}
    C:\mathbb{R}^n &\rightarrow \mathbb{R}^m \\ x&\mapsto\left(\xx\right)_{i\in I},
\end{align*}
its adjoint, the \emph{synthesis operator}
\begin{align*}
    D:\mathbb{R}^m &\rightarrow \mathbb{R}^n \\ 
    (c_i)_{i \in I}&\mapsto \sum_{i\in I} c_i \cdot x_i,
\end{align*}
and the concatenation of analysis, followed by synthesis, the \emph{frame operator}
\begin{align*}
    S:\mathbb{R}^n &\rightarrow \mathbb{R}^n\\
    x&\mapsto \sum_{i\in I} \xx\cdot x_i.
\end{align*}
If $X$ is a frame, then $C$ is injective, $D$ surjective, and $S$ bijective. In $\RR^n$ all the above operators are realized via left-multiplication of $x$ with a corresponding matrix. In this sense, the analysis operator $C$ can be identified with the $m\times n $ matrix
$$C = \begin{pmatrix}
    -  x_1  -\\
%    -  x_2  -\\
      \vdots  \\
    -  x_m  -
\end{pmatrix}.$$
For the synthesis operator, we have that $D=C^\top$. 
Recall that in matrix terminology, injectivity, and surjectivity relate to the corresponding matrix having full rank. Hence, if the weight matrix of a layer in a neural network has full rank, then it can be interpreted as the analysis operator of the frame consisting of its row vectors if $m\geq n$ and as the synthesis operator of the frame consisting of its column vectors if $m\leq n$. At the initialization of a neural network, the weight matrices are commonly set to be Gaussian i.i.d. matrices known to have full rank with probability $1$ \cite{mehta2004random}. Hence, one can be (almost) sure to start the training with the rows, resp. columns of the weight matrices to constitute frames. Here, we concentrate on the case where $m\geq n$ and refer to such a layer as \textit{redundant}.\\
The matrix associated with the frame operator is $S=DC$. It can be used to construct the \emph{canonical dual frame} for $X$, given by $\tilde{X}=\left(S^{-1} x_i\right)_{i\in I}$. Denoting $\tilde{D}$ as the associated synthesis operator leads to the canonical frame decomposition of $x\in \RR^n$ by $X$,
\begin{equation}\label{eq:can}
    x = S^{-1}S = \sum_{i\in I} \xx\cdot S^{-1} x_i= \tilde{D}Cx.
\end{equation}
In this way, \eqref{eq:can} is equivalent to $\tilde{D}$ being a left-inverse of $C$, allowing perfect reconstruction of $x$ from $Cx$.
% If the frame is redundant, there are infinitely many duals, hence left-inverses of $C$.
To reconstruct an input vector from the output of a ReLU-layer, we will construct a left-inverse for it exactly in the spirit of \eqref{eq:can}.\\
Finally, one can find the minimal upper and the maximal lower frame bound in \eqref{eq:frame} via the largest and smallest eigenvalue of $S$ respectively. The ratio $\frac{B}{A}$ of these bounds corresponds to the condition number of the linear mapping given by the analysis operator, hence the weight matrix of the network layer, indicating its numerical stability.

% With the ReLU function given as the piece-wise linear function
% $$\relu(s)=\max(0,s)$$
% for $s\in \RR$,
\subsection{ReLU-layers as Non-linear Analysis Operators}
In a frame-theoretic context, we define the ReLU-layer associated with a collection of vectors $X = \left( x_i \right)_{i\in I} \subseteq \mathbb{R}^n$ and a bias vector $\alpha\in\mathbb{R}^m$ as the non-linear mapping
\begin{align}
\begin{split}
    C_\alpha: \RR^n&\rightarrow \mathbb{R}^m  \\
    x&\mapsto \left( \relu(\langle x,x_i\rangle-\alpha_i ) \right)_{i\in I}.
    % \begin{pmatrix}
    % \relu(\langle x,x_1\rangle-\mathbb{B} )\\
    % \vdots\\
    % \relu(\langle x,x_m\rangle-\mathbb{B}_m ).
    % \end{pmatrix}
\end{split}
\end{align}
The notation $C_\alpha$ is chosen to reflect the link to the frame analysis operator $C$.
Of course, this is equivalent to how a ReLU-layer is commonly denoted, $\relu(Cx-\alpha) $ where ReLU applies component-wise. For fixed $x$, the effect of the shift by the bias $\alpha$ and the ReLU function on the frame analysis can be interpreted as all frame elements with $\xx < \alpha_i$ are set to be the zero-vector. According to this observation, we introduce the notation
\begin{equation}\label{eq:index}
    \I:=\{i\in I:\langle x,x_i\rangle\geq\alpha_i\},
\end{equation}
determining the index set associated with those frame elements which are not affected by the ReLU function for $x$.
%We shall refer to the corresponding elements as ``active'' elements of $X$ for $\alpha$.
This perspective requires referring to sub-collections of frames very often. We write $X_L=(x_i)_{i\in L}$ for the sub-collection of $X$ with respect to the index set $L\subseteq I$. Analogously, we add $L$ as a subscript to the operators associated with $X_L$, e.g. $C_L$ is the analysis operator of $X_L$. Clearly, the case where $L=\I$ plays a central role.

\subsection{Input Data on the Closed Ball}
One of the core ideas in this paper is the restriction of $C_\alpha$ to the closed ball of radius $r>0$ in $\RR^n$, denoted by
$$
\mathbb{B}_r=\{x\in\RR^n:\|x\|\leq r\}.
$$
We write $\mathbb{B}=\mathbb{B}_1$.
Indeed, this is a very reasonable assumption when thinking of standard data normalization practices for neural networks \cite{backprop12,normalization20}. 
It turns out that this restriction allows for a much richer analysis of the injectivity of $C_\alpha$ than on all of $\RR^n$, in particular, involving the radius $r$.
Furthermore, as the output of a ReLU-layer has only non-negative entries, hence lies within $\RR^n_+$, the input domain of any ReLU-layer that applies to the output of a previous ReLU-layer on the ball lies within the non-negative part of $\mathbb{B}_r$, denoted by
\begin{equation}
    \mathbb{B}_r^+ = \mathbb{B}_r\cap\RR^n_{+}.
\end{equation}
%This is due to a trade-off between $\alpha$ and $r$ that one can exploit.
Similarly, we write $\mathbb{B}^+ = \mathbb{B}\cap\RR^n_{+}$.
The boundary of the unit ball, or equivalently, the $(n-1)$-sphere is denoted by
$$
\mathbb{S} = \partial \mathbb{B} = \{x\in\RR^n:\|x\|=1\}.
$$

\section{Injectivity of $C_\alpha$ on $\mathbb{B}_r$}
The ReLU-layer mapping $C_\alpha$ is - by design - non-linear, such that a condition for its injectivity will generally depend on the input. Fixing $x$, one notices that if the sub-collection $X_{\I}$ is a frame, then the analysis operator $C_{\I}$ is injective, which we will use to study the injectivity of $C_\alpha$. For $\alpha \equiv 0$, Puthawala et al. refer to this property as ``$x$ having a directed spanning set'' see Definition 1 in \cite{puth22}.
% \begin{definition}[Directed Spanning Sets, Puthawala et al., 2022]\label{dss}
% Let $W\in \RR^{m\times n}$. We say that $W$ has a directed spanning set (DSS) of $K\subseteq \RR^n$ with respect to a vector $x\in \RR^n$ if there exists a $\hat{W}_x$ such that each row vector of $\hat{W}_x$ is a row vector of $W$,
% \begin{equation}
%     \langle x,w_i \rangle \geq 0 \quad \text{for all } w_i\in \hat{W}_x,
% \end{equation}
% and $K\subseteq \operatorname{span}(\hat{W}_x)$.
% \end{definition}
In the following, we formulate this for general $\alpha$ and $K=\mathbb{B}_r$ in the context of frame theory.  
\begin{definition}[$\alpha$-rectifying on $\mathbb{B}_r$]\label{alpharect}
A collection $X=(x_i)_{i\in I}\subseteq \mathbb{R}^n$ is called $\alpha$-rectifying for $\alpha \in \mathbb{R}^m$ on $\mathbb{B}_r$ if for all $x \in \mathbb{B}_r$ the sub-collection $X_{\I} = (x_i)_{i\in \I}$ is a frame for $\RR^n$. 
\end{definition}
An analogous definition can be formulated for $\mathbb{B}_r^+$. Unless explicitly stated, we always refer to $\mathbb{B}_r$ when writing that $X$ is $\alpha$-rectifying, since it covers the case $\mathbb{B}_r^+$.

In Lemma 2 of the same paper \cite{puth22} the authors show that $\alpha$-rectifying on $\RR^n$ characterizes the injectivity of $\Ta$.
% \begin{lemma}[Injectivity of ReLU-layers on $\RR^n$, Puthawala et al., 2022]
%     Let $W\in \RR^{m\times n}$ and $b\in \RR^m$. The function $\relu(W(\cdot)+b):\RR^n\rightarrow \RR^m$ is injective (on $\RR^n$) if and only if $\relu(W{\vert_{b\geq 0}}(\cdot))$ is injective, where $W{\vert_{b\geq 0}}$ is row-wise the same as $W$ where $b_i \geq0$, and is a row of zeroes when $b_i <0$.
% \end{lemma}
%While the proof of the full characterization for $\mathbb{B}_r$ is beyond the scope of this manuscript, we show one direction for $\mathbb{B}_r^+$, which is also used in the following.
We revisit this characterization for $\mathbb{B}_r$ and $\mathbb{B}_r^+$. 
Again, the frame-theoretic formulation simplifies the statement significantly. % For this contribution, only the implication direction is important. 
\begin{theorem}[Injectivity of ReLU-layers on $\mathbb{B}_r$]\label{reluinj1}
Consider $X=(x_i)_{i\in I}\subseteq \mathbb{R}^n$, $\alpha \in \mathbb{R}^m$. If $X$ is $\alpha$-rectifying on $\mathbb{B}_r$ (resp. $\mathbb{B}_r^+$), then $C_\alpha$ is injective on $\mathbb{B}_r$ (resp. $\mathbb{B}_r^+$).
\end{theorem}
A proof can be found in the appendix.
 % {\xxl , see \cite{xxlehha23}.}
Hence, we can shift the question of injectivity of $\Ta$ to the verification of the $\alpha$-rectifying property for a given collection of vectors $X$.

% \begin{remark}
%     Studying the injectivity of $\Ta$ for arbitrary $K\subseteq \RR^n$ opens a very rich mathematical theory where specific topological properties of $K$ become crucial.
% \end{remark}
\textbf{Stability.}
Following the lines of \cite{bruna14} and again, switching from $\mathbb{R}^n$ to $\mathbb{B}_r$, one can show that the injectivity of $\Ta$ on $\mathbb{B}_r$ implies frame-like inequalities analogous to \eqref{eq:frame}, i.e. 
% Assuming that $\Ta$ is injective on $\mathbb{B}_r$, one can show frame-like inequalities \eqref{eq:frame}, analogous to \cite{bruna14}. 
%it satisfies inequalities, analog to the . In fact, following , 
there are constants $0<A_0\leq B_0<\infty$ such that
\begin{equation}\label{eq:reluframe}
    A_0\cdot \|x\|^2\leq \sum_{i\in I}\left \vert \relu\left( \langle x,x_i \rangle -\alpha_i \right) \right \vert^2 \leq B_0\cdot\| x\|^2
\end{equation}
for all $x\in\mathbb{B}_r$. Here, $A_0$ can be chosen as the smallest eigenvalue and $B_0$ as the largest eigenvalue of all frame operators associated with the frames $X_{\I}$ with $x\in \mathbb{B}_r$.
% A global stability statement as given in \eqref{eq:reluframe} is of theoretical interest, however, from a practical perspective one might be more interested in local statements that reflect the behavior of the ReLU-layer for adversarial examples. {\color{red} ???}

% \begin{theorem}\label{cor:reluinj}
%     Given $X=(x_i)_{i\in I}\subseteq \mathbb{R}^n$, $\alpha \in \mathbb{R}^m$ and $\emptyset\neq K\subseteq \RR^n$ open or strictly convex, then the following are equivalent:
% \begin{enumerate}[(i)]
%     \item $X$ is $\alpha$-rectifying on $K$,
%     \item $C_\alpha$ is injective on $K$.
% \end{enumerate}
% \end{theorem}
\textbf{Inclusiveness.}
It is clear that if $X$ is $\alpha$-rectifying, then $X$ is $\alpha'$-rectifying for all $\alpha'\leq \alpha$. Therefore, we call
\begin{center}
    $\alpha$ an \textit{upper bias} for $\Ta$ if $X$ is $\alpha$-rectifying.  
    %{\color{red} max ???}
\end{center}
This perfectly reflects the role of the bias vector in a neural network: the larger the bias values, the more neurons are activated by the ReLU function, hence the ``more injective'' the ReLU-layer becomes in the sense that it is injective for a larger set of bias vectors. Therefore, it is of natural interest to find the largest possible upper bias for a given weight matrix.
A unique maximal upper bias, however, does not exist in general.
%{\xxl , see \cite{xxlehha23}.}

\textbf{Restriction to $\mathbb{S}$.}
It is important to notice that we may restrict the $\alpha$-rectifying property to unit norm vectors since the norms directly scale the upper bias values $\alpha_i$ and can be re-introduced at any time. In this sense, $X$ is $\alpha$-rectifying if and only if $\overline{X}=\left(x_i \cdot \|x_i\|^{-1}\right)_{i\in I}$  is  $\overline\alpha$-rectifying, where $\overline\alpha_i=\alpha_i \cdot \|x_i \|$.
Therefore, in the following we will always assume $X \subseteq \Sp$, i.e. $\|x_i\|=1$ for all $i\in I$. Note that this corresponds to standard weight normalization \cite{weightnorm16}.

\textbf{Bias-radius interplay.}
Often when studying ReLU-layers theoretically, the bias is implicitly incorporated into the linear part of the operator. However, in our work, we deliberately keep it as a shift as the interplay of bias and input domain is of central interest. We mentioned that an upper bias $\alpha$ favors injectivity when it is large. On the other hand, a large input data domain, i.e. a ball with large radius $r$ offers more flexibility for normalization. However, there is a general trade-off: the larger the radius is chosen, the smaller $\alpha$ will get, in general, and vice versa. We have the following trivial fact:
\begin{center}\label{ex:1}

    Any frame is $\alpha$-rectifying on $\mathbb{B}_r$ for $\alpha \equiv -r$,
    %If it is a basis, (i.e. $m=n$) this is also necessary. 
\end{center}
i.e. $\alpha_i = -r$ for all $i\in I$.
Hence, any redundant ReLU-layer is injective on the closed ball with any radius if the bias vector is sufficiently small.
For a basis, (i.e. $m=n$) the above fact becomes also necessary, immediately implying that a basis can never be $\alpha$-rectifying on $\RR^n$ for any $\alpha$. However, the standard basis for $\RR^n$ is $\alpha$-rectifying on $\mathbb{B}^+$ for $\alpha\equiv 0$.
This shows that taking into account the input domain is a crucial step to take when studying injectivity since it naturally adapts to situations where a frame is not $\alpha$-rectifying on $\RR^n$ but might be on $\mathbb{B}_r$, resp. $\mathbb{B}_r^+$. The question that we are now interested in is, \textit{how} to find a ``good'' upper bias for $\mathbb{B}_r$ and $\mathbb{B}_r^+$?

\begin{figure*}
    \centering
    \includegraphics[width = \linewidth]{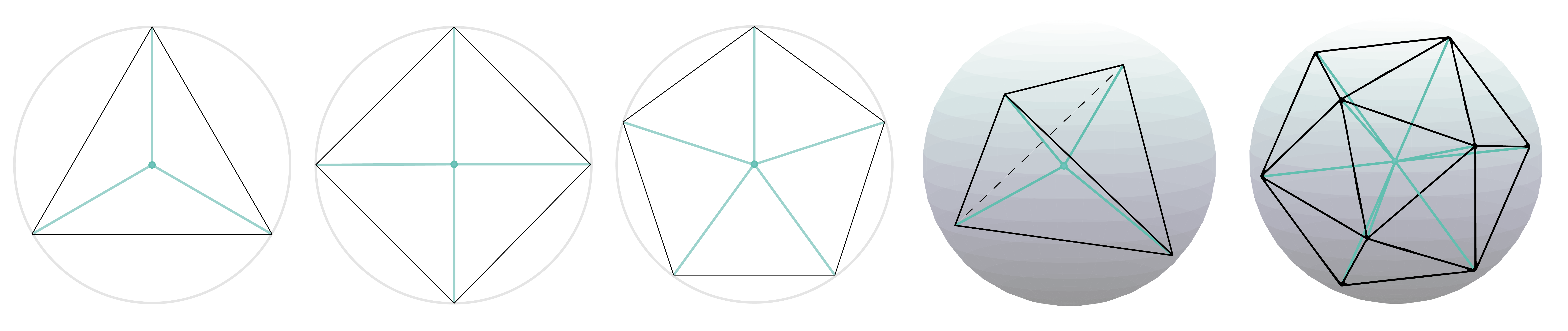}
    \caption{Frame vectors $X$ (blue) and their convex hulls forming convex regular polytopes $P_X$. From left to right: Mercedes-Benz, Square, and Pentagon frame in $\RR^2$, Tetrahedron and Icosahedron frame in $\RR^3$. The unit ball $\mathbb{B}$ is outlined in gray.}
    \label{fig:regular}
\end{figure*}

The Mercedes-Benz frame in $\RR^2$ \cite{casfin12}, given by
    $$
    X_{mb}=\left(
    \begin{pmatrix}
    0 \\
    1 
    \end{pmatrix},
    \begin{pmatrix}
    -\nicefrac{\sqrt{3}}{2} \\
    -\nicefrac{1}{2}
    \end{pmatrix},
    \begin{pmatrix}
    \nicefrac{\sqrt{3}}{2} \\
    -\nicefrac{1}{2}
    \end{pmatrix}
    \right)
    $$
(see Figure \ref{fig:regular}) is a particularly good example, where the optimal upper bias for $\mathbb{B}$ can be found by looking at the geometry of the frame. Its elements determine the vertices of an equilateral triangle so that we can reduce the problem to one pair of elements by symmetry. The worst case is found by $\langle x_i, x_j \rangle=-\frac{1}{2}$. Hence, $X_{mb}$ is $\alpha$-rectifying on $\mathbb{B}$ for $\alpha\equiv-\frac{1}{2}$.
% It is not hard to guess that the redundancy $\frac{m}{n}$ plays a crucial role in the $\alpha$-rectifying property. Already noted by Bruna et al. in \cite{bruna14}, there is the following necessary redundancy condition for $\RR^n$.
% \begin{ex}\label{lem:red}
% A frame that is $\alpha$-rectifying on $\RR^n$ has at least redundancy $2$, i.e. $m\geq 2n$.
% \end{ex}
% One can overcome this so-far-established necessary redundancy $2$ condition by considering a proper subset of $\RR^n$, in our case, $\mathbb{B}_r$ and $\mathbb{B}_r^+$.
This idea can be generalized to polytopes in arbitrary dimensions. In $\RR^3$, we obtain that the Tetrahedron frame, given by
    \begin{equation*}
        X_{tet} = \frac{1}{\sqrt{3}} \cdot \left(\begin{pmatrix}
        1 \\
        1 \\
        1
        \end{pmatrix},
        \begin{pmatrix}
        1 \\
        -1 \\
        -1
        \end{pmatrix},
        \begin{pmatrix}
        -1 \\
        1 \\
        -1
        \end{pmatrix},
        \begin{pmatrix}
        -1 \\
        -1 \\
        1
        \end{pmatrix}\right).
    \end{equation*}
     (see Figure \ref{fig:regular}) is $\alpha$-rectifying on $\mathbb{B}$ for $\alpha \equiv - \frac{1}{\sqrt{3}}$.
    % The icosahedron frame, given by
    % \begin{equation*}
    %     X_{ico} = \frac{1}{\sqrt{1+\varphi^2}}\cdot  \left(\begin{pmatrix}
    %     0 \\
    %     \pm 1 \\
    %     \pm \varphi
    %     \end{pmatrix},
    %     \begin{pmatrix}
    %     \pm 1 \\
    %     \pm \varphi \\
    %     0
    %     \end{pmatrix},
    %     \begin{pmatrix}
    %     \pm \varphi \\
    %     0 \\
    %     \pm 1
    %     \end{pmatrix}\right),
    % \end{equation*}
    % where $\varphi=\frac{1+\sqrt{5}}{2}$ is the golden ratio, is $\alpha$-rectifying on $\mathbb{R}^3\setminus \mathring{B}^n$ for $\alpha \equiv \frac{\varphi}{1+\varphi^2}\approx 0.45$.
%Here, the problem of finding $\alpha$ can be reduced to identifying the smallest analysis coefficients possible for one face of the tetrahedron, since the vectors determining the vertices of the faces form frames (even bases), like for the Mercedes-Benz frame in the $2$-dimensional case.
In a more general setting, where the frame elements are not aligned in a regular manner, we can at least reduce the problem to consider every face individually.%This motivates our approach to find an upper bias, which we formalize in the following.

\section{Convex Polytopes and Bias Estimations}
In a nutshell, we estimate a ``good'' upper bias vector $\alpha$ for a given set of vectors $X$, hence, the ReLU-layer mapping $\Ta$ is injective on $\mathbb{B}_r$. It turns out that the combinatorial structure of the convex polytope associated with the elements of $X$ can be related  to the $\alpha$-rectifying property of $X$. To prepare the estimation procedure, we shall introduce all building blocks for the estimation procedure for $\mathbb{B}_r$ in Section \ref{sec:pbe} and then deduce a version for $\mathbb{B}_r^+$ in Section \ref{relunets}.

For all standard results on convex polytopes, we refer to \cite{zieg12}. Here, we are specifically interested in convex polytopes that arise as the set of all convex linear combinations of a collection of vectors $X=(x_i)_{i\in I}\subseteq\mathbb{S}$,
\begin{equation}\label{eq:poly}
    P_X=\{x\in \RR^n: x = \sum_{i \in I} c_i \cdot x_i, c_i\geq 0, \sum_{i \in I} c_i = 1 \}.
\end{equation}
%This way, it sounds like a fairly natural property, especially when considering a random initialization procedure.
A face of $P_X$ is any intersection of $P_X$ with an affine half-space (in any dimension) such that none of the interior points of $P_X$ (w.r.t. the induced topology on $P_X$) lie on its boundary. While vertices and edges are the $0$- and $1$-dimensional faces of $P_X$, the $(n-1)$-dimensional faces are called \textit{facets}. For every face and, in particular, every facet $F$, there are $a\in \RR^n \setminus \{0\}$ and $b\in \RR$ such that
%this note that one can write a facet equivalently to \eqref{eq:facet1} as a subset of the form
\begin{equation}\label{eq:facet2}
    F=\{x\in P_X:\langle a,x\rangle=b\},
\end{equation}
i.e. any facet lies on an affine subspace of codimension $1$ of $\RR^n$. Furthermore, any $x\in F$ can be written as the convex linear combination, 
\begin{equation*}
x = \sum_{i\in I_{F}}c_i \cdot x_i, \quad c_i\geq 0, \quad \sum_{i\in I_{F}}c_i = 1.
\end{equation*}
We shall write the index set of vertices, associated with $F$ as
\begin{equation}\label{eq:vertex-index}
    I_{F}=\{i\in I:x_i\in F\}.
\end{equation}
The following lemma reveals the core idea of our approach.
\begin{lemma}\label{lem:0}
    Let $F$ be a facet. If $0\notin F$, then $X_{I_{F}}$ is a frame.
\end{lemma}
In other words, as long as the facet does not go through the origin, the associated vertices form a frame. A proof can be found in the appendix.
%For $X\subseteq \Sp$ and $m\geq n+1$, $P_X$ is always full-dimensional, i.e. it does not live on a proper subspace of $\RR^n$. 
%hence
%\begin{equation}\label{eq:facet1}
 %   F=\{x\in \RR^n: x = \sum_{i\in I_{F}}c_i \cdot x_i, c_i\geq 0, \sum_{i\in I_{F}}c_i = 1 \}
%\end{equation}
%any facet is a convex polytope in $\RR^{n-1}$.

We call $X$ \textit{omnidirectional} if $0$ lies in the interior of $P_X$ (w.r.t.~the topology in $\RR^n$), see Definition 1 in \cite{behr18}. Equivalently, there cannot be a hyperplane so that the elements in $X$ are all accumulated on only one side of it.

For the proposed bias estimation on $\mathbb{B}_r$, omnidirectionality is an essential property as it allows to cover every $x\in\mathbb{B}_r$ the same way. For $\mathbb{B}_r^+$ we formulate an analogous condition in Section \ref{relunets}. Moreover, if $X$ is omnidirectional, then $0$ cannot lie on any facet of $P_X$ and Lemma \ref{lem:0} applies. Numerically, it is verified via a simple convex optimization program \cite{behr18}. 

%Furthermore, omnidirectionality is essential to formulate the bias estimation for every $x\in \mathbb{B}_r$.   

Assuming a certain ordering of the facets, we write $F_j$ referring to the $j$-th facets of $P_X$. Analogous to the idea of obtaining the optimal upper biases for the Mercedes-Benz and the Tetrahedron frame, we will use the frames $X_{I_{F_j}}$ for all $j$ to estimate a bias.
Letting the cone of $F_j$ be denoted as
$$
\operatorname{cone}(F_j)=\{tx:x\in F_j,t\geq0\},
$$
then omnidirectionality and $X\subseteq \Sp$ provide the following properties.
\begin{lemma}\label{prop:pol}
If $X\subseteq \Sp$ is omnidirectional, then the following holds.
\begin{enumerate}[(i)]
    \item $\bigcup_{j}I_{F_j}=I$,\label{1}
    \item $\bigcup_j \operatorname{cone}(F_j)=\RR^n$\label{2}
    \item $X_{I_{F_j}}$ is a frame for every $j$.\label{3}
\end{enumerate}
\end{lemma}
These three properties build the backbone of our approach. By \eqref{1}, every frame element is a vertex of $P_X$.
%This property holds since $\Sp$ is the boundary of a convex set.
Due to \eqref{2}, we can partition $\mathbb{B}_r$ into facet-specific conical subsets where we can estimate a bias locally. %This property is a direct consequence of $X$ being omnidirectional.
And most importantly, by \eqref{3}, every sub-collection associated to a facet induces a frame. Properties \eqref{1} and \eqref{2} are easy to see and \eqref{3} is a direct consequence of Lemma \ref{lem:0}.
%(For this property, $0\notin \partial P_X$ already suffices.)
% Altogether, we will identify bias values $\alpha_i$, such that for any choice of $x\in F_j^\mathbb{B}$ with $x_i\in F_j$, there is a facet-sub-frame active, hence $X$ is $\alpha$-rectifying.\\

\begin{remark}
For a facet $F$, the vectors $X_{I_{F}}$ will be redundant ($m>n$) only in rare cases. If the frame elements lie in general position on $\Sp$, then every $X_{I_{F}}$ is a basis ($m=n$) with probability $1$ \cite{randpol}.
\end{remark}

Before we introduce the upper bias estimation procedures for $\mathbb{B}_r$ and $\mathbb{B}_r^+$, we provide an explanation of why the particular grouping of the frame elements into vertices of facets is indeed suitable for the purpose of finding large upper bias values for the $\alpha$-rectifying property. 

If $X$ is omnidirectional and $F$ a facet of $P_X$, then consistent with \eqref{eq:facet2} there are $a\in\mathbb{R}^n\setminus\{0\}$ and $0\neq b\in\mathbb{R}$ such that
\begin{align*}
    \langle a,x_k \rangle& = b,&\text{for } k &\in I_{F},\\
    \langle a,x_\ell \rangle &< b,& \text{for }\ell &\notin I_{F}.
\end{align*}
In this sense, the construction of $X_{I_{F}}$ is a natural way of selecting spanning sub-collections of $X$ with the highest coherence possible, making this particularly useful for our purpose.

\subsection{Polytope Bias Estimation for $\mathbb{B}_r$}\label{sec:pbe}
We now introduce the \emph{Polytope Bias Estimation} (PBE) for $\mathbb{B}_r$ with $r>0$. The procedure estimates an upper bias, denoted as $\alpha^\mathbb{B}$, such that $X$ is $\alpha^\mathbb{B}$-rectifying on $\mathbb{B}$. This implies that $X$ is $(r^{-1}\cdot \alpha^\mathbb{B})$-rectifying on $\mathbb{B}_r$.

The core idea is to partition $\mathbb{B}$ (and $\Sp$) into conical pieces,
\begin{align}
    F_j^\mathbb{B} &:= \operatorname{cone}(F_j)\cap \mathbb{B}\label{eq:cone}\\
    F_j^\mathbb{S} &:= \operatorname{cone}(F_j)\cap \Sp.
\end{align}
If $X$ is omnidirectional, by Lemma \ref{prop:pol}, we have
\begin{align}
    \mathbb{B}=\bigcup_j F_j^\mathbb{B}\quad \text{and}\quad
    \mathbb{S}=\bigcup_j F_j^\mathbb{S}\label{eq:cupB}.
\end{align}
%In other words, for any $x\in \mathbb{B}$ there is $j$ with $x\in F_j^\mathbb{B}$.
To find $\alpha_i^\mathbb{B}$, we identify the minimal analysis coefficient $\yx$ that can occur for $y$ on each $F_j^\mathbb{B}$ containing $x_i$, i.e.
\begin{align}
    \alpha_i^\mathbb{B} & := \min_{
     \substack{y \in F_j^\mathbb{B}\\
    j:x_i\in F_j}
    } \langle y , x_i \rangle\label{eq:ab}.
\end{align}
We do not tackle this optimization problem directly but solve two related problems instead. On the one hand, we consider the minimal auto-correlation values on each facet,
\begin{equation}\label{eq:pbep}
    \alpha_i^X := \min_{
    \substack{\ell \in I_{F_j}\\j:x_i\in F_j}
    } \langle x_{\ell},x_i \rangle,
\end{equation}
that are easy to compute. On the other hand, we solve
\begin{equation}
    \alpha_i^\mathbb{S} :=
        \min_{
        \substack{y \in F_j^\mathbb{S}\\j:x_i\in F_j}
        } \langle y , x_i \rangle\label{eq:pbes}
\end{equation}
via convex linear programs.
Note that the sets, on which all three optimization problems happen are subsets of each other, $F_j^\mathbb{B}\supset F_j^\mathbb{S}\supset X_{I_{F_j}}$, so that we immediately observe that
$\alpha_i^{\mathbb{B}}\leq \alpha_i^{\mathbb{S}}\leq \alpha_i^X$. With this, we solve \eqref{eq:ab}. 
\begin{theorem}(PBE for $\mathbb{B}$)\label{prop:pbe}
    If $X\subseteq \Sp$ is omnidirectional, then $X$ is $\alpha^\mathbb{B}$-rectifying on $\mathbb{B}$ and
    $\alpha_i^\mathbb{B}$, given in \eqref{eq:ab} can be computed as
    \begin{equation}\label{eq:pbeb}
        \alpha_i^\mathbb{B} =
        \begin{cases}
        0 &\text{if } \alpha_i^X \geq 0\\
        \alpha_i^\mathbb{S}\quad &\text{otherwise.}
        \end{cases}
    \end{equation} 
     If $\alpha_i^X<0$, then $\alpha_i^\mathbb{S}$ given in \eqref{eq:pbes} is the minimum over $j:x_i \in F_j$ of the solutions of the convex linear programs
    \begin{align}\label{eq:opt}
        \min\ \left(x_i^\top D_{I_{F_j}}\right)d\nonumber \\
        \text{subject to}\  d &\geq 0\\
        \|D_{I_{F_j}} d\|_2 &\leq 1,\nonumber
    \end{align}
    where $D_{I_{F_j}}$ is the synthesis operator of $X_{I_{F_j}}$.
    %With this, $X$ is $(r^{-1}\cdot \alpha^\mathbb{B})$-rectifying on $\mathbb{B}_r$ for any $r>0$.
\end{theorem}
A proof can be found in the appendix.
The general case follows from $\mathbb{B}_r = \{x\in \RR^n : x = r\cdot y, y\in \mathbb{B}\}$ for $r>0$.

Hence, the minimal argument of \eqref{eq:ab} lies on $\Sp$ or at zero, depending on the sign of the minimal correlation of a facet, given by $\alpha_i^X$.
So, a strategy to obtain $\alpha^\mathbb{B}$ is to start considering the easy-to-compute $\alpha_i^X$ by finding the smallest auto-correlation value with $x_i$ among all facets that are adjacent to $x_i$. Then, only if $\alpha_i^X<0 $, the convex optimization \eqref{eq:opt} has to be solved. See Algorithm 1 for a pseudo-code of the procedure.

\begin{figure}[t]
    \centering
    \includegraphics[width = \columnwidth]{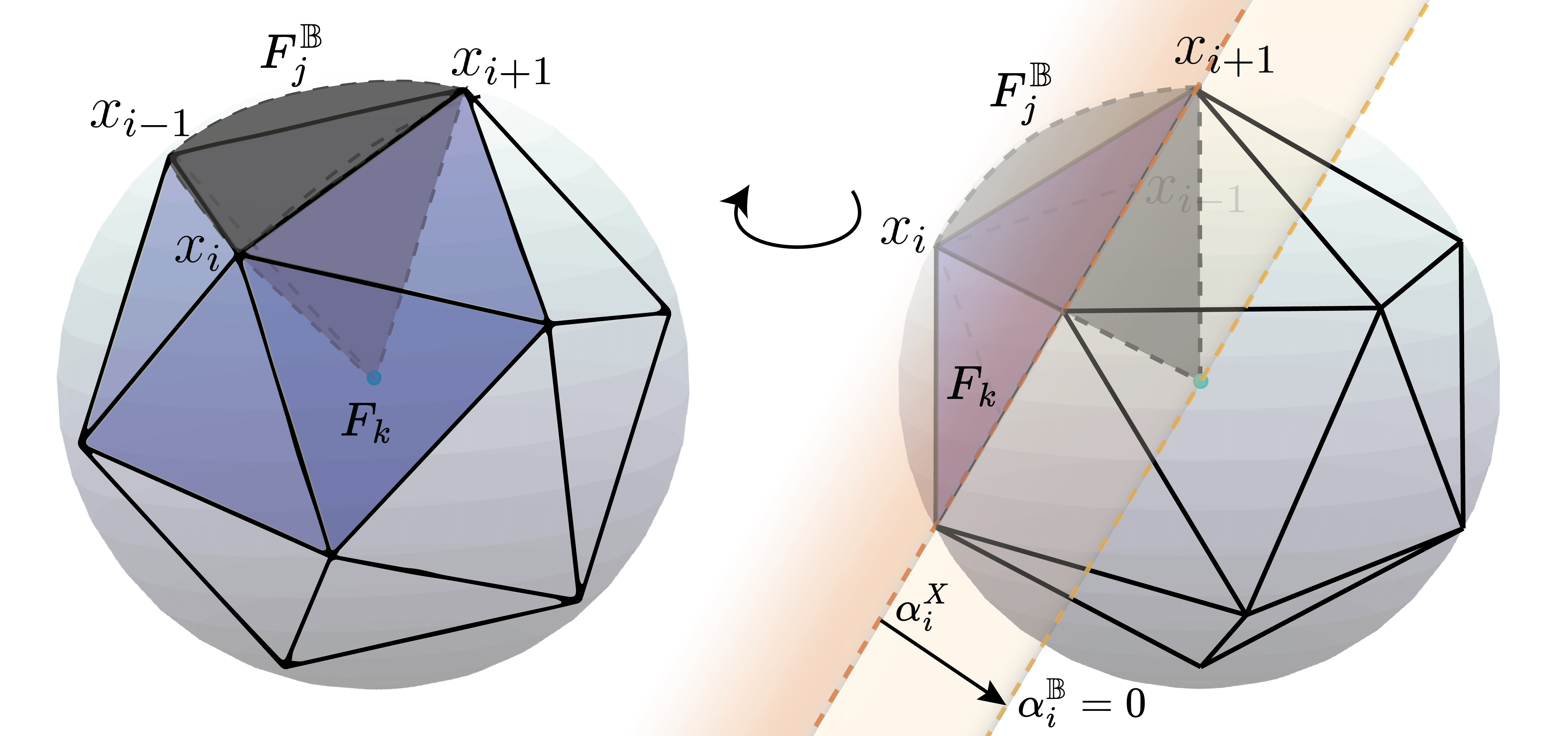}
    \vspace{-0.2cm}
    \caption{Geometrical intuition of the PBE for the Icosahedron frame on $\mathbb{B}$.
    Consider $x_i$.
    Left: The (blue) filled facets are used to compute $\alpha_i^\mathbb{B}$. The (gray) darker piece (dashed border) indicates $F_j^\mathbb{B}$. Right: Rotated perspective of the left image. The affine half-space $\Omega_i = \{x\in \RR^n: \xx \geq \alpha_i^X, i\in I_{F_j} \}$, indicated by the left-most area with decreasing opacity (orange) contains all vectors such that all vertices of the adjacent facets are active. Since $\alpha_i^X\geq 0$, the brighter (yellow) half-space represents the solution $\alpha_i^\mathbb{B} = 0$.} 
    \label{fig:pbe}
\end{figure}

\begin{ex} (a) For the Tetrahedron frame $X_{tet}$, we have
        $\alpha^X \equiv - \frac{1}{3}$, therefore $\alpha^\mathbb{B} = \alpha^\mathbb{S} \equiv -\frac{1}{\sqrt{3}}.$ \\
(b) For the Icosahedron frame, given by
    \begin{equation*}
        X_{ico} = \frac{1}{\sqrt{1+\varphi^2}}\cdot  \left(\begin{pmatrix}
        0 \\
        \pm 1 \\
        \pm \varphi
        \end{pmatrix},
        \begin{pmatrix}
        \pm 1 \\
        \pm \varphi \\
        0
        \end{pmatrix},
        \begin{pmatrix}
        \pm \varphi \\
        0 \\
        \pm 1
        \end{pmatrix}\right),
    \end{equation*}
    (see Figure \ref{fig:regular}), where $\varphi=\frac{1+\sqrt{5}}{2}$ is the golden ratio,
    % is $\alpha$-rectifying on $\mathbb{R}^3\setminus \mathring{B}^n$ for $\alpha \equiv \frac{\varphi}{1+\varphi^2}\approx 0.45$.
    we have $\alpha^X \equiv \frac{\varphi}{1+\varphi^2}\approx 0.45$, therefore $\alpha^\mathbb{B}\equiv 0$. Figure \ref{fig:pbe} shows the idea of the PBE for this example geometrically.
\end{ex}

Note that $0\geq \alpha^\mathbb{B}_i$ is reasonable to guarantee the $\alpha$-rectifying property on $\mathbb{B}_r$ since for any upper bias $\alpha$ and $x=0$, it has to hold that $\langle 0,x_i \rangle = 0\geq\alpha_i$ for all $i$ in some $I_{F_j}$.
% Note that computing $\min_{\substack{y \in F_j^\mathbb{S}\\j:x_i\in F_j}} \langle y , x_i \rangle$ in this way requires solving as many convex optimization programs as there are adjacent facets to $x_i$.

\subsection{Polytope Bias Estimation for $\mathbb{B}_r^+$}\label{relunets}
In neural networks, often ReLU-layers succeed each other. In this context, we show that $\mathbb{B}_r^+$ is conceptually the right input domain for a PBE for ReLU-layers that are applied to the output of a previous one. In fact, this requires knowing where the image of $\mathbb{B}_r$ under $\Ta$ lies. 
\begin{lemma}\label{lem:image}
    Let $X$ be $\alpha$-rectifying and $B_0$ denote the largest optimal upper frame bound among $X_{\I}$ with $x\in \mathbb{B}_r$. Then
    %and denote the image of $\mathbb{B}_r$ under $\Ta$ as
    %$$
    %C_\alpha\left(\mathbb{B}_r\right)=\{z\in \RR^m: z = \Ta x, x \in \mathbb{B}_r\}.
    %$$
    %Then
    \begin{equation}\label{eq:estim}
        C_\alpha\left(\mathbb{B}_r\right) \subseteq \mathbb{B}_{r\sqrt{B_0}}^+.
    \end{equation}
\end{lemma}
It is easy to show that \eqref{eq:estim} is a direct consequence of the upper inequality in \eqref{eq:reluframe} and clearly, holds for $x\in\mathbb{B}_r^+$ as well. Note that we may also estimate the radius of the ball as $r\sqrt{B}$, where $B$ is any upper frame bound of $X$.
%Rephrasing the proposition, the image of the ball with radius $r>0$ under $C_\alpha$ lies in the non-negative part of the ball with radius $r\sqrt{B_0}$.
%Hence, $\mathbb{B}_{r\sqrt{B_0}}^+$ is a suitable domain to compute the PBE for a subsequent ReLU-layer.

\begin{figure}[tb]
    \centering
    \includegraphics[width = \columnwidth]{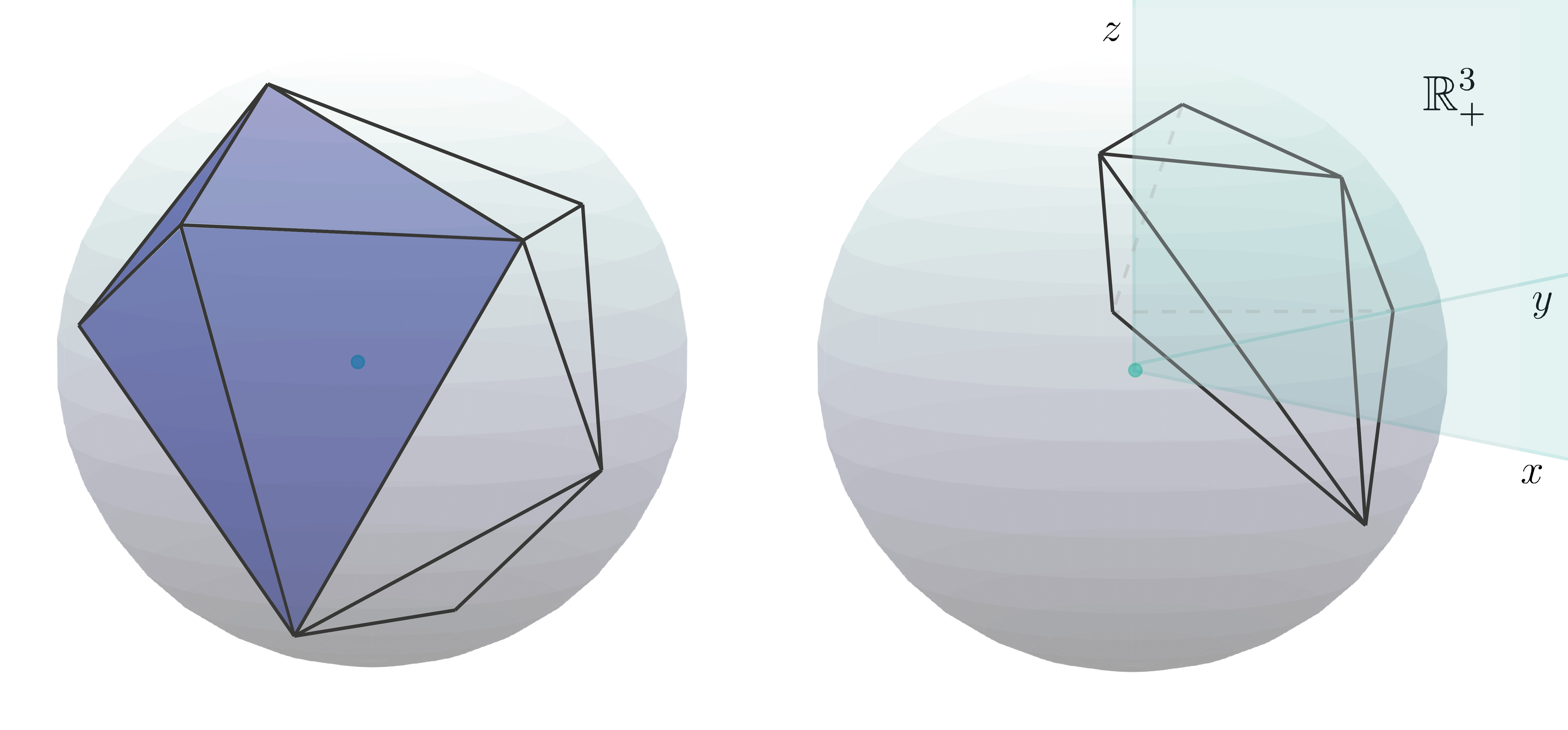}
    \vspace{-0.2cm}
    \caption{Non-regular polytopes. Left: 
    %If there are spread-out faces on $P_X$, the estimation will naturally become coarser (smaller). 
    %{\bf Unclear. Maybe:} 
    The estimated bias values $\alpha_i^\mathbb{B}$ are computed from the largest adjacent facet of $x_i$. Hence, the less regular the normalized frame elements are distributed on the sphere, the smaller $\alpha^\mathbb{B}$ becomes.
   % if the polytope is more regular - then more elements pf $P_X$ are 'above' the hype-plane  the estimation will naturally become coarser (smaller). } 
    % Yet, this is in concordance with how the true bias $\alpha$ would behave as well. 
%     {\xxl \bf Unclear. } 
    Right: The frame is non-negatively omnidirectional since $\bigcup_{j\in J^+} F_j \supseteq \RR^n_+$, but not omnidirectional.}
    \label{fig:irreg}
\end{figure}

We approach the PBE for $\mathbb{B}_r^+$ by restricting the computations of the PBE introduced in Theorem \ref{prop:pbe} to only those facets, that actively contribute to the estimation. In this sense, we only consider those frame elements whose associated facets have a non-trivial intersection with $\RR^n_+$. We denote the corresponding index sets as
$$
J^+=\{j : F_j\cap \RR^n_+\neq \emptyset\} ,\qquad I^+ = \bigcup_{j\in J^+} I_{F_j}.
$$
According to this, instead of omnidirectionality, we only have to require
\begin{align}
    %\quad \text{and}\quad
    \RR^n_+&\subseteq\bigcup_{j\in J^+} \operatorname{cone}(F_j), \quad \text{and}\label{eq:posomn1}\\
    0&\notin F_j\ \text{ for all } j\in J^+,\label{eq:posomn2}
\end{align}
which we shall refer to as \textit{non-negative omnidirectionality}. See Figure \ref{fig:irreg} (right) for an illustration. This is tailored to provide the properties in Lemma \ref{prop:pol} for $\mathbb{B}_r^+$: by \eqref{eq:posomn1}, we have analogously to \eqref{eq:cupB},
\begin{equation}\label{eq:cupB+}
    \mathbb{B}^+ \subseteq \bigcup_{j\in J^+} F_j^\mathbb{B}
\end{equation}
and condition \eqref{eq:posomn2} is sufficient for $X_{I_{F_j}}$ being a frame for every $j\in J^+$ by Lemma \ref{lem:0}. With this, we have all requirements to deduce the PBE for $\mathbb{B}_r^+$.

\begin{theorem}[PBE for $\mathbb{B}^+$]\label{prop:pbe+}
    If $X\subseteq \Sp$ is non-negatively omnidirectional, then $X$ is $\alpha^{\mathbb{B}^+}$-rectifying on ${\mathbb{B}^+}$ with
    \begin{equation}\label{eq:a+}
        \alpha_i^{\mathbb{B}^+}=
        \begin{cases}
            \alpha_i^\mathbb{B} &\text{ for } i \in I^+\\
            s_i &\text{ else,} 
        \end{cases}
    \end{equation}
    % \begin{equation}
    %     \alpha_i^{B^+}= \min_{
    %     \substack{y \in F_j^\mathbb{B}\\
    %     j\in J:x_i\in F_j \cap X^+}
    %     } \langle y , x_i \rangle,
    % \end{equation}
    where $s_i\in \RR$ is arbitrary.
\end{theorem}
A proof can be founded in the appendix. This reduces the computational cost and improves the upper bias estimation as potentially large bias values in $I \setminus I^+$ can be omitted.

% \begin{ex}
%     Let $X=\left(\begin{pmatrix}
%         1 \\
%         0
%         \end{pmatrix},
%         \begin{pmatrix}
%         0 \\
%         1
%         \end{pmatrix},
%         \begin{pmatrix}
%         -\nicefrac{1}{\sqrt{2}} \\
%         -\nicefrac{1}{\sqrt{2}}
%         \end{pmatrix}\right)$ and $\alpha \equiv 0$. Then $X$ is not $\alpha$-rectifying on $\mathbb{B}$ but is on $\mathbb{B}^+$ since $x_3$ is omitted in the PBE for $\mathbb{B}_r^+$.
% \end{ex}

\begin{remark}\label{rem:+}
    Clearly, conditions \eqref{eq:posomn1} and \eqref{eq:posomn2} are weaker than omnidirectionality, yet, harder to check numerically. Similarly, $F_j\cap \RR^n_+ \neq \emptyset$ is not straightforward to verify. Indeed, it holds true for all adjacent facets of $x_i\in \RR^n_{+}$, however, there might be facets meeting the condition but with no vertices in $\RR^n_+$. The interested reader will find a continued discussion in the appendix. Finding an efficient implementation of this, however, is left as an open problem.
\end{remark}

\subsection{Remarks on the Optimality of the PBE}
In general, we cannot expect the proposed PBE to yield upper biases that are maximal. Estimating the error of the estimation to a maximal upper bias (if exists), however, is difficult since this would require knowing the combinatorial structure (i.e. the vertex-facet index sets) of a general polytope, which has been a topic of active research for several decades. In the special cases when the polytopes are regular and simplicial (every facet has exactly $n$ vertices), e.g. Mercedes-Benz, Tetrahedron and Icosahedron frame, we expect that the estimated upper bias is indeed maximal. It is easy to verify that the PBE is also stable to perturbations as long as the combinatorial structure is preserved. Hence, one could expect that the estimation will be more accurate the more evenly distributed the frame elements are on the sphere. See Figure \ref{fig:irreg} (left) for an illustration.

\begin{algorithm}[tb]
    \caption{PBE for $\mathbb{B}_r$}
   \label{alg:pbe}
   \begin{algorithmic}
%\STATE Normalize $x_i \leftarrow \frac{x_i}{\|x_i\|}$
\STATE Get $I_{F_j}$ via computing $V_X$
\FOR{$j=1, \dots, J$}
    \STATE $\beta_j = \min_{k<\ell\in I_{F_j} } \langle x_k,x_\ell\rangle$
    \ENDFOR
    \FOR{$i=1,\dots,m$}
        \STATE $\alpha_i^* = \min\limits_{j \text{ s.t. } i\in I_{F_j}} \beta_j$
        \IF{$\alpha_i^* \geq 0$} 
                    \STATE $\alpha_i^\mathbb{B} \gets 0$
                \ELSE
                    \STATE $y^* \gets $ solution of \eqref{eq:opt}
                    \STATE $\alpha_i^\mathbb{B} \gets y^* \cdot r^{-1}$
        \ENDIF
    \ENDFOR
\end{algorithmic}
\end{algorithm}

\subsection{Local Reconstruction via Facets}
Unless $\I \neq I$ for all $x\in \mathbb{B}_r$, there cannot be only \textit{one} global left-inverse for $\Ta$. We propose to systematically construct a collection of left-inverses, each associated with one facet of $P_X$.
%Assuming omnidirectionality, for every $x$ there is $j$ such that $x\in F_j^\mathbb{B}$ can be reconstructed from $z = \Ta x$ using the $j$-th left-inverse.
Recall that the frame operator of $X_{I_{F_j}}$ is denoted by $S_{I_{F_j}}$ and that its canonical dual frame is given by
$
\Tilde{X}_{I_{F_j}}=\left(S_{I_{F_j}}^{-1}x_i\right)_{i\in I_{F_j}}.
$
%The grouping of the frame elements into facet-sub-frames gives rise to a very natural approach to performing reconstruction from

%Under the assumptions of Theorem \ref{pro:pbe2} this naturally extends to the decomposition of $\left( P_{X}^{\circ}\right)^{\mathrm{C}}$ into $\operatorname{cone}^+(F_j)$. For now, we refer to both situations when we write $F_j$ in the following.\\
\begin{theorem}\label{rec}
Let $X\in \Sp$ be $\alpha$-rectifying on $\mathbb{B}$ and omnidirectional.
For every $x\in \mathbb{B}$ there is $j$ such that
\begin{equation}\label{eq:leftinv}
    \Tilde{D}_{I_{F_j}}\Ta x = x,
\end{equation}
where
%$x\in F_j^\mathbb{B}$,
\begin{align}\label{eq:can-non}
\begin{split}
    \Tilde{D}_{I_{F_j}}:\mathbb{R}^m &\rightarrow \mathbb{R}^n\\
    (c_i)_{i \in I}&\mapsto \sum_{i\in I_{F_j}}  \left(c_i+\alpha_i\right)\cdot S^{-1}_{I_{F_j}} x_i.
\end{split}    
\end{align}
%is a left-inverse of $\Ta$.
\end{theorem}
In other words, $\Tilde{D}_{I_{F_j}}$ is a left-inverse of $\Ta$ for all $x\in F_j^\mathbb{B}.$
By \eqref{eq:cupB}, every $x\in \mathbb{B}$ lies in some $F_j^\mathbb{B}$, hence indeed for any $x\in \mathbb{B}$ there is a left-inverse.
It is easy to see that \eqref{eq:leftinv} reduces to the usual canonical frame decomposition \eqref{eq:can} of $x$ by $X_{I_{F_j}}$.
%Clearly, $\Tilde{X}_{I_{F_j}}$ is in general worse conditioned than the canonical dual of $X_{\I}$, however, the facet-specific duals can be used for a much larger and nicer subset of $\mathbb{B}$.
%

\begin{remark}
    In general, there are infinitely many duals for $X$ \cite{ole}. The canonical dual mentioned above relates to the pseudo-inverse of the associated analysis operator, hence induces the optimal inverse by means of ridge regression.
\end{remark}

\begin{algorithm}[tb]
\caption{Reconstruction via Facets}
\label{alg:rec}
\begin{algorithmic}
    \STATE Get $I_{F_j}$ via computing $V_X$
    \FOR{$j=1,\dots,J$}
        %\STATE $K\gets I_{F_j}$
        \STATE $S^{-1}_{I_{F_j}}\gets\left( (C_{I_{F_j}})^\top C_{I_{F_j}} \right)^{-1}$
        \STATE $\overline{X}\gets X_{I_{F_j}}$
        \STATE $\Tilde{D}_{I_{F_j}} \gets
            \begin{pmatrix}
                \vert & \vert &  & \vert \\
                S^{-1}_{I_{F_j}} \overline{x}_1 & S^{-1}_{I_{F_j}} \overline{x}_2 & \cdots & S^{-1}_{I_{F_j}} \overline{x}_{\vert I_{F_j} \vert}\\
                \vert & \vert &  & \vert
            \end{pmatrix}$
        \STATE 
    \ENDFOR
    \STATE $z= \Ta x$
    \STATE $\overline{z} \gets z+\alpha $
    \WHILE{$j=1,\dots,J$}
        \IF{$I_{F_j}\in \I$}
            \STATE $\Tilde{D}_{I_{F_j}}\overline{z}_{\vert_{I_{F_j}} }=x$
        \ENDIF
    \ENDWHILE
\end{algorithmic}

% \begin{enumerate}[(1)]
%     \item Get $\Io$ from $z_0=C_\alpha x_0$
%     \item Choose $J\in \Io$ s.t. $X_J$ is a frame\\ (e.g. $J = \Io$ or $J=I_{F_{k_0}}$ for some facet $F_{k_0}$)
%     \item Compute $S^{-1}_J=\left( D_J C_J \right)^{-1}$
%     \item Build the matrix $$\Tilde{D}_J =
%             \begin{pmatrix}
%                 \vert & \vert &  & \vert\\
%                 S^{-1}_J x_1 & S^{-1}_J x_1 & \cdots & S^{-1}_J x_m\\
%                 \vert & \vert &  & \vert
%             \end{pmatrix}$$
%     \item Apply $\Tilde{D}_J$ to ${z_0}_{\vert_J }+\alpha_{\vert_J}$ (restriction to $J$)
% \end{enumerate}

\end{algorithm}

\subsection{Implementation}\label{chap:num}
We discuss the implementation aspects of the PBE for $\mathbb{B}$ and the reconstruction formulas.
Our implementations of the algorithms are publicly available under \href{https://github.com/danedane-haider/Alpha-rectifying-frames}{https://github.com/danedane-haider/Alpha-rectifying-frames}.

\subsubsection{PBE}
The vertex-facet index sets $I_{F_j}$ are encoded in what is called the \textit{vertex-facet incidence matrix} $V_X$. Assuming that $P_X$ has $J$ facets, then $V_X$ is the $J\times m$ matrix with entries
$$
V_X[j,i] = \begin{cases}
    1\quad &\text{ if } i\in I_{F_j}\\
    0 &\text{ else,}
\end{cases}
$$
indicating which vertices correspond to which facets.
To compute the vertex-facets incidences, we use the routine \texttt{VERTICES\_IN\_FACETS} from the open-source software Polymake \cite{polymake}. This routine requires the vertices in homogeneous coordinates, i.e.
$$ C_{hom} = 
    \begin{pmatrix}
        1 - x_1 - \\
 %       1 - x_2 - \\
        \vdots \\
        1 - x_m -
    \end{pmatrix}.
$$
%The output is a list containing all $I_{F_j}$. 
Already noted in \cite{puth22}, checking the $\alpha$-rectifying property is probably NP-hard.
%might only be feasible with an exponential time algorithm.
The computation of $V_X$ relies on convex-hull algorithms, which are also not expected to run in polynomial time for general polytopes. However, for points in general position on $\Sp$ (i.e. no hyperplane in $\RR^n$ contains more than $n$ of the points), the ``reverse-search'' algorithm is expected to finish in linear time in the number of vertices $m$ for fixed dimension $n$ \cite{convhull}. This condition is precisely fulfilled when assuming random initialization and normalized frame elements. Polymake uses this algorithm via the command \texttt{prefer "lrs";}.

Algorithm \ref{alg:pbe} gives step-by-step instructions to compute $\alpha^\mathbb{B}$ for any omnidirectional frame $X\subseteq \Sp$.

\subsubsection{Reconstruction}
In practice, one can read off $\I$ from $z = \Ta x$ and find a facet $F_{j}$ such that $I_{F_{j}}\subseteq \I$ using the vertex-facet incidences. Note that $F_{j}$ might not be unique with this property.

Algorithm \ref{alg:rec} describes how the systematic construction of the left-inverses $\Tilde{D}_{I_{F_j}}$ can be done, assuming that $X$ is $\alpha$-rectifying and omnidirectional.
%and how any $x\in \mathbb{B}$ can be reconstructed from $z = \Ta x$.

\section{Numerical Experiments}\label{sec:num}
A series of experiments revealed that the injectivity behavior of a ReLU-layer is very sensitive to many hyperparameters and circumstances, such as the size of the layer, the depth of the network, the position of the layer within the network, initialization and normalization procedures, the optimizer and the data itself.
Here, we present a numerical experiment, where we want to focus merely on the size of the ReLU-layer, i.e. its redundancy. Therefore, the experimental setting is designed to be as simple and reduced as possible.
Considering more realistic network models requires a much broader study, which goes beyond the scope of this contribution.

\begin{figure}[t]
    \centering
    \includegraphics[width = \columnwidth]{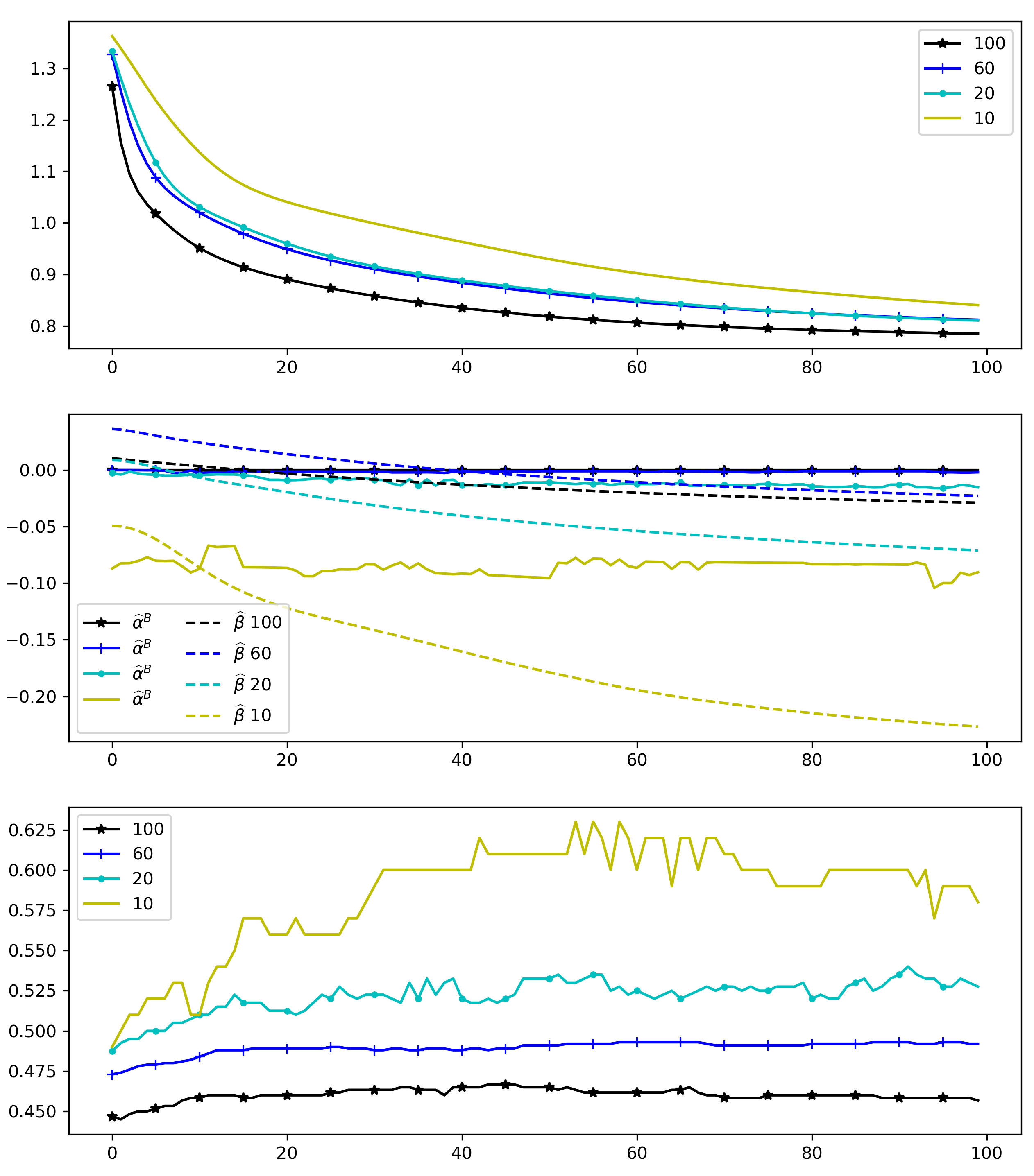}
    \caption{Averaged quantities (across $10$ iterations) related to a ReLU-layer over $100$ epochs of training with different redundancies $m=\vert I\vert$. Top: Cross entropy loss on the validation set. Mid: Mean of the trained biases, $\widehat{\beta}$ (dashed), and mean of the estimated upper biases on $\mathbb{B}_r$, $\widehat \alpha^\mathbb{B}$ (solid). Bottom: Proportion of learned bias values that are smaller than the estimations, i.e. $\#(\beta_i \leq \alpha_i^\mathbb{B})/m$, indicating the injectivity trend.}
    \label{fig:plot}
\end{figure}

\subsection{Experimental Setting}
We train a neural network with one ReLU-layer and a soft-max output layer on the Iris data set \cite{iris36}. For the ReLU-layer, we consider four redundancy settings $m=\vert I\vert=10,20,60,100$. The corresponding networks are mappings from $\RR^4 \rightarrow \RR^m \rightarrow \RR^3$. After normalization to zero mean and a variance of one, all data samples lie within the ball of radius $r=3.1$. We use the upper biases from the PBE for $\mathbb{B}$ (Theorem \ref{prop:pbe}) with appropriate scaling to monitor the injectivity behavior of the ReLU-layers during training, see Figure \ref{fig:plot}.
Here, optimization of a cross-entropy loss is done using stochastic gradient descent at a learning rate of $0.5$ for $100$ epochs.

\subsection{Discussion}
The top plot shows that in our setting, high redundancy in the ReLU-layer yields the smallest validation loss. Expectedly, high redundancy also increases the chance of the polytope $P_X$ having many non-negatively correlated facets, i.e. $\alpha_i^X\geq 0$. Hence, more bias estimations are $0$ according to \eqref{eq:pbeb}. The mid-plot shows this nicely (solid lines).
%In this sense, a very redundant ReLU-layer has better chances to be injective.
Note that all learned biases $\beta$ decrease in mean (dashed lines). The lower the redundancy, the stronger this decrease is. Since the bias estimations $\alpha_i^\mathbb{B}$ remain almost unchanged in mean, we may conclude that lacking injectivity of low redundancy ReLU-layers (i.e. too many output values are zero) is compensated during training via the bias.
%relative to the estimations $\hat{\alpha}$, most noticeable for low redundancies.
The bottom plot shows another measure of the injectivity trend: the proportion of learned bias values smaller than the estimations, i.e. $\#(\beta_i \leq \alpha_i^\mathbb{B})/m$. An increase in this quantity indicates that a ReLU-layer is becoming ``more injective'' during training. In concordance with the previous observations, layers with low redundancy show a stronger increase here as well. This could be interpreted as high redundancy favors injectivity from the start.
%This is an indication of a stronger trend towards injectivity, caused by a strong decrease in the learned biases, relative to the estimations.

In this sense, the PBE can help us to better understand the role of injectivity in neural networks. In the example, we are able to see the effect of different sizes of a ReLU-layer in regard to injectivity and validation loss and, in particular, what happens when the layer is chosen too small.
%Interpreting this behavior seems difficult: While the biases decrease (in mean) for the sake of minimizing the loss function, the geometry of the ReLU-layer appears to stay almost unchanged (in mean). Hence, the ReLU-layers introduce less sparsity, underlining the trend towards injectivity.
%Note that in the proposed setting, the ReLU-layers were not verified by the PBE to be injective on $\mathbb{B}_r$.
It remains an open question if these results are representative for other settings and which are the responsible components causing this behavior.
Future numerical investigation is necessary for a better understanding.

\section{Conclusion}
We presented a frame-theoretic setting to study the injectivity of a ReLU-layer on the closed ball with radius $r>0$ in $\RR^n$ and on its non-negative part. Moreover, we introduced a systematic approach of verifying it in practice, called polytope bias estimation (PBE). This method exploits the convex geometry of the weight matrix associated with a ReLU-layer and estimates a bias vector such that the layer is injective on the ball for all biases smaller or equal to the estimation. This allows us to give sufficient and quantified conditions for the invertibility of a ReLU-layer. Corresponding reconstruction formulas are provided.
%Restricting the algorithms to frames with an omnidirectionality property is convenient and also reasonable, yet, we believe that a weakening of this condition is possible.
Via a straightforward implementation, the PBE allows to study the injectivity behavior of a redundant ReLU-layer and perform perfect reconstruction of the layer input where applicable. So far, our work contributes to a better understanding of the behavior of neural network layers by means of observations without interaction with the actual optimization procedure. As a possible application, the estimated upper biases from the PBE could be used to design a regularization procedure where the bias is guided toward injectivity during training.

\section*{Acknowledgment}
D. Haider is recipient of a DOC Fellowship of the Austrian Academy of Sciences at the Acoustics Research Institute (A 26355). The work of M. Ehler was supported by the WWTF project CHARMED (VRG12-009) and P. Balazs was supported by the OeAW Innovation grant project FUn (IF\_2019\_24\_Fun), and the FWF projects LoFT (P 34624) and NoMASP (P 34922). The authors would like to thank Daniel Freeman for fruitful and fun discussions and Lukas Köhldorfer for his valuable feedback.

\bibliographystyle{icml2023}
\bibliography{references}

%%%%%%%%%%%%%%%%%%%%%%%%%%%%%%%%%%%%%%%%%%%%%%%%%%%%%%%%%%%%%%%%%%%%%%%%%%%%%%%
%%%%%%%%%%%%%%%%%%%%%%%%%%%%%%%%%%%%%%%%%%%%%%%%%%%%%%%%%%%%%%%%%%%%%%%%%%%%%%%
% APPENDIX
%%%%%%%%%%%%%%%%%%%%%%%%%%%%%%%%%%%%%%%%%%%%%%%%%%%%%%%%%%%%%%%%%%%%%%%%%%%%%%%
%%%%%%%%%%%%%%%%%%%%%%%%%%%%%%%%%%%%%%%%%%%%%%%%%%%%%%%%%%%%%%%%%%%%%%%%%%%%%%%
\newpage
\appendix
\onecolumn

\section{Proofs}

\subsection{Theorem \ref{reluinj1}}
% Following \cite{puth22}.
\begin{proof}
Assume $C_\alpha x = C_\alpha y$, for $x,y\in \mathbb{B}_r$. Clearly,
\begin{equation}\label{eq:000}
\xx > \alpha_i\quad \Leftrightarrow\quad \yx > \alpha_i,
\end{equation}
so that we derive 
\begin{equation}\label{eq:xy etc}
\langle x,x_i\rangle = \langle y,x_i\rangle,\quad \text{ for all }i \in \I\cap\Iy.
\end{equation}
Since $\mathbb{B}_r$ is convex, for $\lambda \in (0,1)$ it holds that $x_\lambda := (1-\lambda)x + \lambda y \in \mathbb{B}_r$. 
For $i\in I_{x_\lambda}^\alpha$, we compute 
$$
\langle x_\lambda,x_i\rangle = (1-\lambda)\langle x,x_i\rangle+\lambda\langle y,x_i\rangle. 
$$
Since $\langle x_\lambda,x_i\rangle\geq\alpha_i$, at least one of the two, $\langle x,x_i\rangle$ or $\langle y,x_i\rangle$, must be bigger or equals $\alpha_i$. Without loss of generality, let us suppose that $\langle x,x_i\rangle\geq \alpha_i$. If we have $\langle x,x_i\rangle>\alpha_i$, then \eqref{eq:000} leads to $\langle y,x_i\rangle>\alpha_i$. If $\langle x,x_i\rangle=\alpha_i$, then 
$\langle x_\lambda,x_i\rangle\geq\alpha_i$ implies $\langle y,x_i\rangle=\alpha_i$. Thus, we have verified that 
% Due to \eqref{eq:000}, we obtain $\langle x,x_i\rangle, \langle y,x_i\rangle>\alpha_i$. If $\langle x_\lambda,x_i\rangle=\alpha_i$, then  
% by the linearity of the inner product and \eqref{eq:000},
% \begin{align}
%     \langle x_\lambda , x_i \rangle &> \alpha_i\quad \Rightarrow \quad \xx > \alpha_i,\ \yx > \alpha_i\label{eq:del1}
% \end{align}
% and
% \begin{align}
%     \langle x_\lambda , x_i \rangle &= \alpha_i\quad \Rightarrow\quad \xx = \alpha_i,\  \yx = \alpha_i.\label{eq:del2}
% \end{align}
$I^\alpha_{x_\lambda}\subseteq \I\cap \Iy$. Since  $x_\lambda\in \mathbb{B}_r$, the $\alpha$-rectifying property yields that $X_{I^\alpha_{x_\lambda}}$ is a frame. Hence, $X_{\I\cap \Iy}$ is a frame. According to \eqref{eq:xy etc}, we deduce $x=y$. Therefore, $\Ta$ is injective on $\mathbb{B}_r$.

Since $\mathbb{B}_r^+$ is a convex set as well, the statement for $\mathbb{B}_r^+$ can be proven analogously.
\end{proof}

\subsection{Lemma \ref{lem:0}}

We recall that a collection of points $p_1,\dots,p_m$ in $\RR^n$ are called \textit{affinely independent} if $\sum_{i=1}^m \lambda_i p_i = 0$ and $\sum_{i=1}^m\lambda_i = 0$ imply that $\lambda_1=...=\lambda_m=0$.

\begin{proof}
    It is a known fact that any facet of a polytope in $\RR^n$ contains at least $n$ affinely independent vertices \cite{zieg12}. Let $F$ be a facet and w.l.o.g. $x_1,...,x_n\in X_{I_F}$ be affinely independent. If $0\notin F$, by \eqref{eq:facet2} there is $a\in \RR^n \setminus \{0\}$ and $b\in \RR$ with $b\neq 0$ such that
    \begin{equation}\label{eq:aff}
        \langle a, x_i \rangle = b, \qquad i=1,...,n.
    \end{equation}
    Assuming $\sum_{i=1}^n \lambda_i x_i = 0$ and using \eqref{eq:aff}, from
    \begin{align}
        0 = \langle a, \sum_{i=1}^n \lambda_i x_i \rangle = \sum_{i=1}^n \lambda_i \langle a, x_i \rangle = \sum_{i=1}^n \lambda_i b,
    \end{align}  
    it follows that $\sum_{i=1}^n \lambda_i = 0$. Since $x_1,...,x_n$ are affinely independent, $\lambda_1=...=\lambda_n=0$. Therefore, $x_1,...,x_n$ are linearly independent, hence $X_{I_F}$ is a frame for $\RR^n$.
\end{proof}

\subsection{Theorem \ref{prop:pbe}}

\begin{proof}
    We first show that if $\alpha_i^X\geq 0$, then $\alpha_i^\mathbb{S}=\alpha_i^X$.
    
    By omnidirectionality, $\bigcup_j F_j^\mathbb{S} = \Sp$. Hence, for every $y\in \Sp$ there is $x_y$ in some facet $F_j$ such that $y= \frac{x_y}{\|x_y\|}\in F_j^\mathbb{S}$. We use that $x_y$ can be written as a convex combination of elements of $X_{I_{F_j}}$, i.e. $x_y = \sum_{\ell\in I_{F_j}} c_\ell x_\ell$ with $c_\ell \geq 0$ for all $ \ell\in I_{F_j}$ and $\sum_{\ell\in I_{F_j}} c_\ell = 1$. Hence,
        %y = \frac{x_y}{\|x_y\|} = \sum_{\ell\in I_{F_j}} \underbrace{c_\ell\frac{1}{\|x_y\|}}_{=:d_\ell}  x_\ell,
    \begin{equation}\label{eq:conv}
        y = \frac{x_y}{\|x_y\|} = \sum_{\ell\in I_{F_j}} \frac{c_\ell}{\|x_y\|} x_\ell = \sum_{\ell\in I_{F_j}} d_\ell x_\ell.
    \end{equation}
    Let $\alpha_i^X\geq 0$, i.e. $\langle x_{\ell},x_i \rangle\geq 0$ for all $\ell \in I_{F_j}$ and  $j:x_i\in F_j$. Using \eqref{eq:conv} with $d_\ell = \frac{c_\ell}{\|x_y\|}\geq 0$ and $\sum_{\ell\in I_{F_j}} d_\ell \geq 1 $, we obtain $ \langle y,x_i \rangle \geq \min_{\ell \in I_{F_j}} \langle x_\ell,x_i \rangle$. Thus, we derive 
    \begin{equation}\label{eq:SX}
    \alpha_i^\mathbb{S}= \min_{
    \substack{y \in F_j^\mathbb{S}\\j:x_i\in F_j}
    } \langle y,x_i \rangle \geq
    %\min_{
    %\substack{(d_\ell)_{\ell \in I_{F_j}}: \sum_{\ell\in I_{F_j}} d_\ell x_\ell \in F_j^\mathbb{S} \\j:x_i\in F_j}
    %} \sum_{\ell\in I_{F_j}} d_\ell \langle x_\ell,x_i \rangle \geq 
    \min_{
        \substack{\ell \in I_{F_j}\\j:x_i\in F_j}
        } \langle x_{\ell},x_i \rangle = \alpha_i^X.    
    \end{equation}
    Recalling that $\alpha_i^X\geq \alpha_i^\mathbb{S}$ for all $i \in I$ yields the claim.
    
    Next we show \eqref{eq:pbeb},
    % Let $x_i \in F_j$ then for all $x\in F_j^\mathbb{S}$ it holds that
    % $$
    % \xx \geq \min_{\ell\in I_{F_j}} \langle x_{\ell},x_{i}\rangle
    % $$
    % if this minimum is non-negative. Otherwise,
    % $$
    % \xx \geq \min_{y \in F_j^\mathbb{S}} \langle y , x_i \rangle.
    % $$
    % By taking the minima over $j:x_i\in F_j$ we obtain that $X$ is get the required estimation, analog to \eqref{eq:finest}.
    % Since $(x_i)_{i\in I_{F_j}}$ is a frame for $\mathbb{R}^n$, $X$ is $\alpha$-rectifying on $S^{n-1}$.\\
    using that $F_j^\mathbb{B}=\{y\in \mathbb{B}:y=\frac{x}{t},\exists x\in F_j^\mathbb{S}, \exists t\in [1,\infty)\}$ and $\langle x,x_i \rangle \geq \alpha_i^{\mathbb{S}}$ for all $x\in F_j^{\mathbb{S}}$ with $x_i\in F_j$.
    
    Let $\alpha_i^X\geq 0$, then $\alpha_i^\mathbb{S}\geq 0$ by \eqref{eq:SX}. Hence $\xx\geq 0$ for all $x\in F_j^\mathbb{S}$. We deduce,
    $$\alpha_i^\mathbb{B}=\min_{
     \substack{y \in F_j^\mathbb{B}\\
    j:x_i\in F_j}
    } \langle y , x_i \rangle = \min_{
     \substack{x \in F_j^\mathbb{S}, t\geq 1\\
    j:x_i\in F_j}
    } \langle \frac{x}{t} , x_i \rangle = 0.$$
    %$$
    %\langle \frac{x}{t}, x_i \rangle \geq \frac{1}{t} \alpha_i^\mathbb{S}\rightarrow 0,
    %$$    
    %hence, $\alpha_i^{\mathbb{B}} = 0$.
    Now let $\alpha_i^X < 0$. For $x\in F_j^{\mathbb{S}}$ with $x_i\in F_j$, we distinguish two cases. If $\xx \geq 0$, then $\langle \frac{x}{t}, x_i \rangle \geq 0$. If $\xx < 0$, then $\langle \frac{x}{t}, x_i \rangle \geq \xx \geq \alpha_i^\mathbb{S}$. Since $\alpha_i^{\mathbb{S}}\leq \alpha_i^X < 0$, we deduce $\alpha_i^{\mathbb{B}}\geq \alpha_i^{\mathbb{S}}$. 
    % $\xx < 0$, then $\langle \frac{x}{t}, x_i \rangle \geq \xx \geq \alpha_i^\mathbb{S}$
    % \begin{enumerate}[ ]
    %     \item $\xx \geq 0$, then $\langle \frac{x}{t}, x_i \rangle \geq 0$, hence  $\alpha_i^{\mathbb{B}} \geq 0$.
    %     \item $\xx < 0$, then $\langle \frac{x}{t}, x_i \rangle \geq \xx \geq \alpha_i^\mathbb{S}$, hence $\alpha_i^{\mathbb{B}} \geq \alpha_i^\mathbb{S}$. 
    % \end{enumerate}
    Recalling that $\alpha_i^\mathbb{B}\leq \alpha_i^\mathbb{S}$ for all $i \in I$ we obtain $\alpha_i^{\mathbb{B}} = \alpha_i^\mathbb{S}$.
    % Together, we have for every $x\in F_j^\mathbb{S}$,
    % \begin{equation*}\label{eq:fint}
    %     \langle \frac{x}{t}, x_i \rangle \geq \min\left( 0, \alpha_i^\mathbb{S} \right),\qquad \text{for all } t\in [1,\infty),
    % \end{equation*}

    To see that $X$ is $\alpha^\mathbb{B}$-rectifying on $\mathbb{B}$, take an arbitrary $z\in \mathbb{B}$. By \eqref{eq:cupB}, there is $j$ such that $z\in F_j^\mathbb{B}$.
    Hence, $\langle z,x_i \rangle\geq \alpha_i^\mathbb{B}$ holds for any $i\in I_{F_j}$. In other words, $I_{F_j}\subseteq I_z^{\alpha^\mathbb{B}}$. Since $X_{I_{F_j}}$ is a frame by $(iii)$ in Lemma \ref{prop:pol}, $X_{I_z^{\alpha^\mathbb{B}}}$ is also a frame, showing the claim.

    Finally, recalling that $D_{I_{F_j}}$ is the synthesis operator of $X_{I_{F_j}}$ we may rewrite Equation \eqref{eq:conv} for $y\in F_j^{\mathbb{S}}$ as $y = \sum_{\ell\in I_{F_j}} d_\ell x_\ell=D_{I_{F_j}}d.$ Then $\min_{\substack{y \in F_j^\mathbb{S}}} \langle y , x_i \rangle$ for $i\in I_{F_j}$ can be formulated as the linear program
    \begin{align*}
        \min\ \left(x_i^\top D_{I_{F_j}}\right)d \\
        \text{subject to}\  d &\geq 0\\
        \|D_{I_{F_j}} d\|_2 &= 1.
    \end{align*}
    If $\alpha_i^X<0$, then the above minimum is negative since $\alpha_i^\mathbb{S}\leq\alpha_i^X<0$. Hence, we can replace $\|D_{I_{F_j}} d\|_2=1$ by $\|D_{I_{F_j}} d\|_2\leq 1$ making the problem convex.
\end{proof}

%\subsection{Lemma \ref{prop:outside}}
%
%\begin{proof}
%    Since $X_\varepsilon$ has the same combinatorial structure as $X$ and is still omnidirectional, we can write
%    $$
%    \RR^n \setminus \mathring B_\varepsilon = \{ y\in \RR^n : y = t\varepsilon x, x\in \Sp, t \geq 1 \}.
%    $$
 %   The statement follows from $\langle t \varepsilon x,x_i \rangle  \geq \varepsilon \alpha_i^{S}$ for $t\geq 1$ and $\varepsilon \geq 0$.
%\end{proof}
%
%\subsection{Lemma \ref{lem:image}}
%
%\begin{proof}
%    By \eqref{reluframe}, $\|\Ta x\|^2 \leq B_0 \|x\|^2 \leq B_0 \cdot r^2$ for $x\in %\mathbb{B}_r$.
%\end{proof}

\subsection{Theorem \ref{prop:pbe+}}
\begin{proof}
    Let $z\in \mathbb{B}^+$.
    By \eqref{eq:cupB+}, there is $j\in J^+$ such that $z\in F_j^\mathbb{B}$. Since $F_j\cap \RR^n_+\neq \emptyset$ we have in particular that $I_{F_j}\subseteq I^+$.
    Hence, analog to the proof of Theorem \ref{prop:pbe}, $\langle z,x_i \rangle\geq \alpha_i^\mathbb{B}$ holds for any $i\in I_{F_j}$. With $\alpha^{\mathbb{B}^+}$ defined as in \eqref{eq:a+} we have that $I_{F_j}\subseteq I_z^{\alpha^\mathbb{B}}\subseteq I_z^{\alpha^{\mathbb{B}^+}}$. Since $X_{I_{F_j}}$ is a frame by Lemma \ref{lem:0}, $X_{I_z^{\alpha^{\mathbb{B}^+}}}$ is also a frame.
\end{proof}

\section{Remarks}

\subsection{Remark \ref{rem:+}}

For fixed $j$, one may verify $F_j\cap \RR^n_+\neq \emptyset$ via the feasibility of the convex optimization problem
\begin{align}\label{eq:opt+}
   \min\ \|D_{I_{F_j}}c\|_2 \nonumber \\
   \text{subject to}\ c&\geq 0\\
   \sum_i c_i &= 1.\nonumber
\end{align}
Indeed, if \eqref{eq:opt+} has a solution, then there is $c\in \RR^n_+$ that can be written as a convex linear combination of the vertices of $F_j$, hence, $F_j\cap \RR^n_+\neq \emptyset$. We suggest the following strategy: Label all facets with vertices in $\RR^n_+$ and continue solving \eqref{eq:opt+} for all adjacent facets. If there is no vertex in $\RR^n_+$ at all, solve \eqref{eq:opt+} for the facets that contain vertices $x_k$ with only small negative components.

\end{document}